\documentclass[pdflatex,sn-mathphys-num]{sn-jnl}% Math and Physical Sciences Numbered Reference Style
\usepackage{graphicx}%
\usepackage{multirow}%
\usepackage{amsmath,amssymb,amsfonts}%
\usepackage{amsthm}%
\usepackage{mathrsfs}%
\usepackage[title]{appendix}%
\usepackage{xcolor}%
\usepackage{textcomp}%
\usepackage{manyfoot}%
\usepackage{booktabs}%
\usepackage{algorithm}%
\usepackage{algorithmicx}%
\usepackage{algpseudocode}%
\usepackage{listings}%
\usepackage{tikz-cd}
\theoremstyle{thmstyleone}%
\theoremstyle{thmstyletwo}%

\theoremstyle{thmstylethree}%

\begin{document}

\title[Large Language Models as a Semantic Interface and Ethical Mediator in Neuro-Digital Ecosystems: Conceptual Foundations and a Regulatory Imperative]{Large Language Models as a Semantic Interface and Ethical Mediator in Neuro-Digital Ecosystems: Conceptual Foundations and a Regulatory Imperative}


\author[1]{\fnm{Alexander V.} \sur{Shenderuk-Zhidkov}}\email{ a.shenderyuk@yandex.ru}
\equalcont{These authors contributed equally to this work.}

\author*[2,3]{\fnm{Alexander E.} \sur{Hramov}}\email{hramovae@gmail.com}
\equalcont{These authors contributed equally to this work.}

\affil[1]{\orgname{Immanuel Kant Baltic Federal University}, \orgaddress{\street{14 A. Nevskogo street}, \city{Kaliningrad}, \postcode{236041}, \country{Russia}}}

\affil*[2]{\orgname{Plekhanov Russian University of Economics}, \orgaddress{\street{36 Stremyanny Lane}, \city{Moscow}, \postcode{115054},  \country{Russia}}}

\affil[3]{\orgname{Pirogov National Medical and Surgical Center}, \orgaddress{\street{70 Nizhnyaya Pervomayskaya street}, \city{Moscow}, \postcode{105203},  \country{Russia}}}


\abstract{This article introduces and substantiates the concept of Neuro-Linguistic Integration (NLI), a novel paradigm for human-technology interaction where Large Language Models (LLMs) act as a key semantic interface between raw neural data and their social application. We analyse the dual nature of LLMs in this role: as tools that augment human capabilities in communication, medicine, and education, and as sources of unprecedented ethical risks to mental autonomy and neurorights. By synthesizing insights from AI ethics, neuroethics, and the philosophy of technology, the article critiques the inherent limitations of LLMs as semantic mediators, highlighting core challenges such as the erosion of agency in translation, threats to mental integrity through precision semantic suggestion, and the emergence of a new `neuro-linguistic divide' as a form of biosemantic inequality. Moving beyond a critique of existing regulatory models (e.g., GDPR, EU AI Act), which fail to address the dynamic, meaning-making processes of NLI, we propose a foundational framework for proactive governance. This framework is built on the principles of Semantic Transparency, Mental Informed Consent, and Agency Preservation, supported by practical tools such as NLI-specific ethics sandboxes, bias-aware certification of LLMs, and legal recognition of the neuro-linguistic inference. The article argues for the development of a `second-order neuroethics,' focused not merely on neural data protection but on the ethics of AI-mediated semantic interpretation itself, thereby providing a crucial conceptual basis for steering the responsible development of neuro-digital ecosystems.}

\keywords{
Large Language Models,
Neuroethics,
Semantic Mediation,
Mental Autonomy,
Neurorights, Neuro-Linguistic Integration}



\maketitle

\section{Introduction: From Synergy to Symbiosis---The Emergence of a New Paradigm}\label{sec1}

The rapid development of neurotechnologies and artificial intelligence (AI), which in recent years has been primarily viewed through the paradigm of their synergy or co-evolution \cite{yuste2017four, gong2025ai, Shenderyuk-Zhidkov2025Harmonization, Shenderyuk-Zhidkov2026Co}, has opened new horizons for medicine, rehabilitation, and the study of brain cognitive functions. However, the accelerated evolution of AI technologies has led to a qualitative leap beyond mere complementarity. The emergence and widespread adoption of powerful Large Language Models (LLMs) signifies the formation of a new semantic layer of interaction between humans and machines \cite{myers2024foundation, mahowald2024dissociating}. LLMs are no longer merely text-processing tools; they are becoming universal interpreters and generators of meaning, capable of operating with context, cultural codes, and social conventions \cite{bharathi2024analysis}. It is this potential that radically transforms the nature of AI integration with neurotechnologies, shifting it from the plane of technical synergy to the realm of deep symbiosis, where the boundary between neural processes and their linguistic expression is mediated by AI.

This transition gives rise to a fundamentally new and still underexplored problem. Whereas classical brain-computer interfaces (BCIs) addressed the task of directly decoding signals into commands \cite{hramov2021physical}, integration with LLMs transforms it into a task of semantic interpretation and context-dependent generation \cite{liu2025talking}. The model does not simply `read' an activity pattern associated with an intention to raise a hand; it attempts to reconstruct and verbalize a thought, emotion, or complex desire \cite{jiang2025cognitive}. This `animation' of a digital intermediary creates an extensive `blind spot' within existing ethical and legal frameworks, which were formulated for the era of data, expert systems, and algorithms, but not for the era of semantic mediators like LLMs, which possess the appearance of understanding and agency.

The aim of this article is to conceptualize this emerging phenomenon, which we define as Neuro-Linguistic Integration (NLI), and to provide a systematic analysis of the transformations it engenders. We seek to demonstrate how the agentic capabilities of LLMs radically reshape the traditional ethical challenges of neurotechnology: data confidentiality acquires the dimension of thought process confidentiality, personal autonomy faces the threat of semantic manipulation, and the problem of access justice is exacerbated by the risk of a `neuro-linguistic divide'---a new form of social inequality based on differential access to technologies for the semantic interpretation of neural data and, consequently, to the quality of cognitive enhancement, communication, and medical diagnosis.
The scientific novelty of this work lies in the synthesis of approaches from AI ethics, neuroethics, and the philosophy of technology to develop the conceptual foundations necessary for formulating adequate and anticipatory regulatory strategies in a context where technology is encroaching not only upon the sphere of private behavior but also upon the very fabric of meaning-making.

\section{Research Methodology}\label{sec2}
This study is grounded in a comprehensive interdisciplinary methodology designed to conceptualize the phenomenon of NLI and analyze the resulting ethical and legal challenges. The uniqueness of the research object---the symbiosis of LLMs and neurotechnologies---demands a synthesis of approaches capable of encompassing its technological specificity, philosophical-ethical implications, and normative-regulatory contexts. The methodological framework comprises four complementary components.

\subsection{Philosophical-Ethical Analysis}
This serves as the foundation for identifying and conceptualizing fundamental problems arising at the intersection of human consciousness and semantically agentic AI. It employs:
\begin{itemize}
\item	Conceptual analysis of key notions such as `mental autonomy,' `neurorights,' `agency,' and `semantic mediator' to clarify their meaning within the new technological reality.
\item	Normative-ethical reflection, drawing on the principles of bioethics (respect for autonomy, non-maleficence, justice) and contemporary AI ethics (explainability, accountability, human-centeredness) to assess how these principles are transformed under the influence of NLI.
\item	A phenomenological approach to describe changes in the structure of subjective experience and personal identity of a user whose intentions and internal states are processed through the interpretive filter of an LLM.
\end{itemize}
\subsection{Conceptual Modelling}
This method is applied to structure the subject area and construct theoretical models. It involves:
\begin{itemize}
\item	Developing an architectural model of an NLI system, defining the role and functions of the LLM as a semantic interface between neural data and social contexts (medicine, communication, education).
\item	Constructing classifications (typologies) of risks and ethical contradictions characteristic of NLI. This methodological step involves identifying and categorizing specific conflicts arising in neuro-linguistic integration systems. An illustrative example is the `authenticity vs. coherence' dilemma in neurocommunication, which involves a conflict between the accurate, albeit imperfect, transmission of the user's original intention and the generation of a stylistically flawless but potentially distorted utterance. Creating such classifications serves as the basis for systematizing and structuring the entire problem field of the research, enabling a shift from describing individual cases to analyzing stable patterns.
\item	Creating conceptual maps as an analytical tool for visualizing and analyzing the interrelationships between LLM capabilities, emerging ethical dilemmas, and potential regulatory measures. This method allows for the clear identification of cause-effect chains (e.g., how the function of `semantic interpretation of neural patterns' generates the risk of `loss of authenticity,' necessitating the principle of `semantic transparency') and for systematizing the complex problem field of NLI.
\end{itemize}
\subsection{Comparative Analysis of Regulatory Approaches}
This analysis aims to assess the adequacy of existing and emerging legal frameworks in addressing the challenges of NLI. The focus is on:
\begin{itemize}
\item	Key legislating jurisdictions: The European Union (the AI Act, the General Data Protection Regulation, GDPR), the United States (sectoral regulation and executive branch initiatives), and China (Administrative Provisions on Generative AI). The selection is based on their active role in shaping global trends and the diversity of their regulatory paradigms---from stringent risk-based approaches to flexible sectoral models.
\item	Analysis criteria: The ability of legal norms to account for the specificity of neural data as the foundation of mental life; to regulate not only the collection/processing of data but also the process of its semantic interpretation by AI; and to allocate responsibility within hybrid human-machine decision-making systems.
\item	Identifying gaps and formulating recommendations for adapting regulatory tools (such as `ethics sandboxes,' certification, and the principle of informed consent) to the realities of NLI.
\end{itemize}
\subsection{Critical Literature Review}
This provides the theoretical and factual foundation for the research. It is conducted at the intersection of three discursive fields:
\begin{itemize}
\item	AI Ethics: Analyzing works dedicated to algorithmic bias, explainability, fairness, and fundamental principles of AI development \cite{floridi2018ai4people, mittelstadt2016ethics}.
\item	Neuroethics: Examining literature on issues of neural data privacy, cognitive enhancement, neurorights, and autonomy in the context of neurotechnologies \cite{ienca2017towards, yuste2017four, kellmeyer2020ethical,  friedrich2021clinical,  botes2022brain}.
\item	Philosophy of Technology: Engaging with concepts that describe the mediating role of technology in human experience, the transformation of agency, and the emergence of hybrid systems (e.g., post-phenomenology, mediation theory) \cite{verbeek2024moralizing,  wood2025postphenomenology}. Recent works analyzing the anthropological impact of LLMs and their role in reshaping cognition and communication are also incorporated \cite{bender2021dangers, shanahan2024talking, heersmink2024phenomenology}.
\end{itemize}

This article is not descriptive but adopts a critical-synthetic character, which allows for the identification of not only points of convergence among these discourses but also their `blind spots,' specifically related to the semantic power of LLMs.
The integration of these methods proceeds sequentially: from the philosophical-ethical elaboration of basic concepts and the modelling of NLI architecture, through critical literature review, to the evaluation of regulatory landscapes. The result is the formation of a holistic conceptual basis for understanding NLI as a new technological and anthropological paradigm and for developing principles for its responsible governance. This approach overcomes the fragmentation inherent in traditional analyses and offers a systemic vision of the challenges at the nexus of neuroscience, AI, and the humanities.

\section{The Technological Paradigm: LLMs as a Semantic Interface in NLI}\label{sec3}

The technological core of the emerging NLI paradigm is the fundamentally new role of Large Language Models (LLMs), transforming them from text-processing tools into a meaning-making interface between biological neural activity and digital ecosystems. This transition marks the overcoming of limitations inherent in previous stages of brain-computer interaction and opens up unprecedented opportunities alongside novel categories of technological risk.

\subsection{From Signal to Meaning: LLM as a Semantic Translator}\label{subsec3.1}

The evolution of BCIs has progressed from decoding simple binary commands using data from non-invasive neuroimaging (e.g., EEG, MEG, fNIRS, fMRI) or invasive sensors based on implantable electrodes---such as motor imagery for command selection \cite{guo2024novel} or P300 evoked potential registration for letter mental selection \cite{chaudhary2025noninvasive}---towards attempts to recognize more complex intentions and patterns related to cognitive states \cite{chaudhary2016brain,  hramov2021physical, pisarchik2025passive}. However, traditional machine learning algorithms used in BCIs face a fundamental barrier: they operate on signal correlations but are incapable of their semantic interpretation within a broad social and personal context. It is precisely this barrier that is overcome by integrating BCIs with LLMs. The number of studies dedicated to this new possibility has been growing rapidly since 2025 \cite{caria2025integrating, caria2025towards}, with applications already emerging in areas like smart manufacturing and Industry 4.0 \cite{asgher20convergence}. This convergence marks a shift from decoding prespecified commands to interpreting open-ended cognitive and linguistic states, a direction anticipated in recent neuroengineering literature (e.g., \cite{kellmeyer2020ethical}).

The LLM can act as a decisive component, performing the translation of neurosemantic patterns (recognized and classified by the BCI) into coherent language, meaningful actions, and contextually relevant responses. If the BCI answers the question, `Which motor area did the user activate?', the LLM addresses the task: `What does the user want to say or do, given this activation, the current situation, their history, and cultural norms?'. For instance, an ambiguous neural pattern associated with frustration might be interpreted by the LLM as a need for help, a desire to change tasks, or an expression of physical discomfort---depending on the context derived from prior interactions. Determining the user's true intention with certainty is impossible and requires their feedback for further model refinement. Thus, LLMs become a semantic bridge, converting neurophysiological data into a subjectively meaningful and socially acceptable output --- raising novel concerns about authenticity and agency that move beyond traditional BCI ethics frameworks \cite{burwell2017ethical}. This process, however, inherently introduces a new layer of mediation where the user's raw mental state is filtered through the statistical priors and cultural embeddings of the LLM, raising immediate questions about interpretative fidelity and the preservation of authentic intent---core neuroethical concerns for the NLI paradigm.

\subsection{Architecture of the NLI System: The Triadic Role of the LLM}\label{subsubsec3.2}

\begin{figure}[ht!]
\centering
\includegraphics[width=1.0\textwidth]{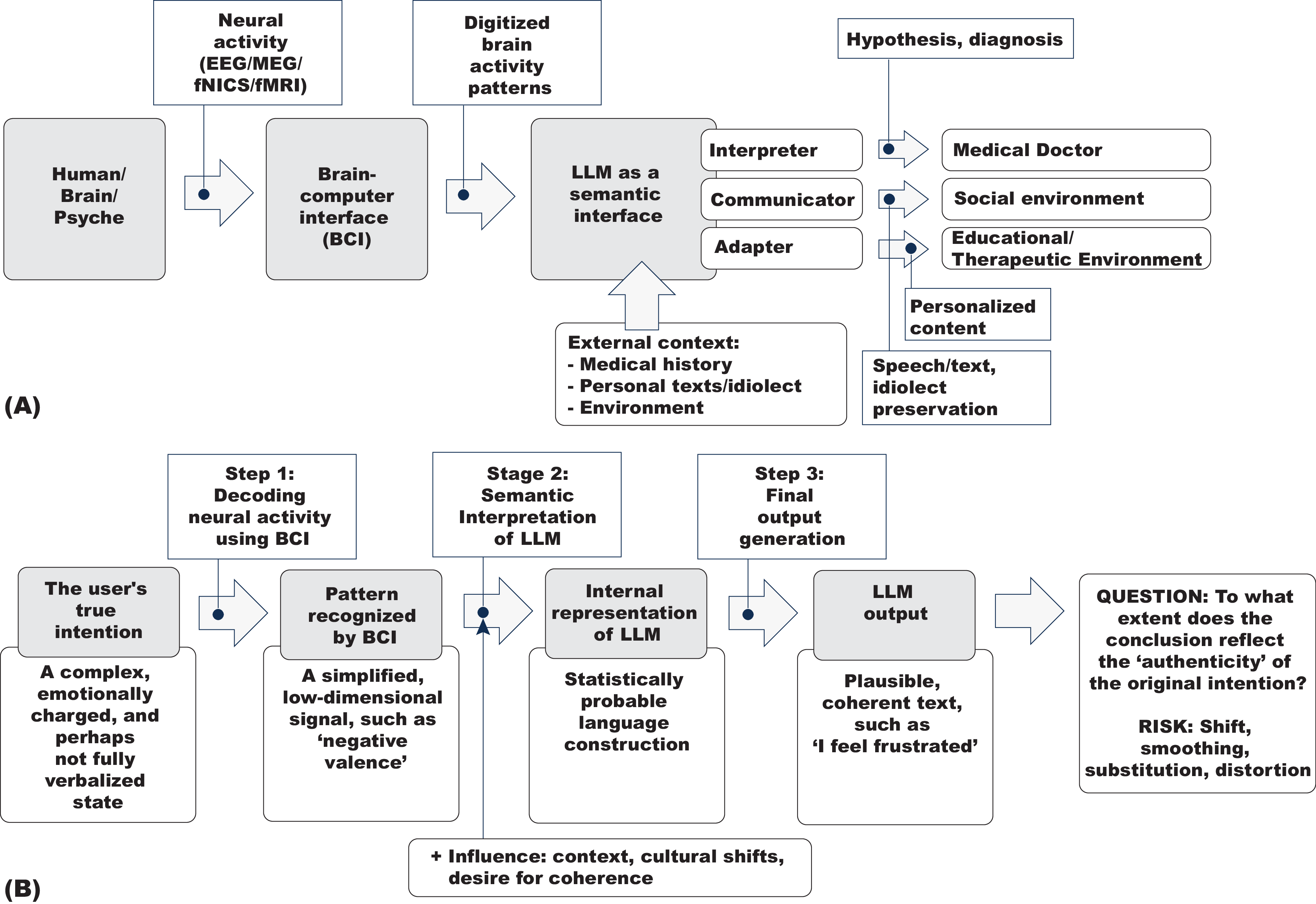}
\caption{\textbf{Conceptual model of NLI and the phenomenon of agency erosion.}
\textbf{(A)} The tri-level architecture of an NLI system, illustrating the role of the Large Language Model (LLM) as the central semantic interface. The schema depicts the transformation of biological neural activity (Level 1: BCI decoding) into formalized patterns, which the LLM (Level 2: the semantic interface) enriches with external context (medical history, personal texts, environment). At Level 3, the LLM executes three differentiated functions depending on the target environment: the Interpreter (generates diagnostic hypotheses for medicine), the Communicator (creates a speech avatar for social interaction), and the Adapter (personalizes educational or therapeutic content based on neurofeedback). This forms an open, iterative loop between the user and the digital ecosystem.
\textbf{(B)} The schema of sequential transformation and potential distortion of the user's original intention (`agency erosion in translation'). The process comprises: (1) BCI-stage reduction: the user's rich internal state is reduced to a simplified signal (e.g., a binary code), leading to a loss of semantic nuance. (2) LLM semantic interpretation: the received signal is filtered through the model's algorithmic priors, cultural biases from training data, and a drive for coherence, forming a probabilistic linguistic construct. (3) Final output generation: the internal construct is rendered as a stylistically polished text or command, the correspondence of which to the user's authentic intention becomes problematic. The diagram illustrates a fundamental risk of NLI---the substitution of authentic expression with an optimized yet potentially alien semantic simulation, thereby questioning the preservation of mental sovereignty in hybrid cognitive systems. This visual model underscores the core neuroethical challenges of interpretative fidelity and the safeguarding of agential authenticity in semantically mediated human-AI symbiosis.
}
\label{fig1}
\end{figure}

The conceptual architecture of a Neuro-Linguistic Integration system, illustrated in Figure \ref{fig1}(A), presents a three-level model where the Large Language Model (LLM) functions as the central semantic interface, transforming biological neural activity into socially meaningful actions and interpretations. This architecture fundamentally diverges from traditional BCIs through the introduction of a context-dependent, meaning-forming layer.

Level 1: From Biological Activity to Digital Patterns. The initial stage (Human (brain/mind) $\rightarrow$ BCI) corresponds to the classic brain-computer interface paradigm. Neural activity, recorded via methods such as EEG, fMRI, or invasive sensors, is digitized and converted into formalized patterns. The key limitation at this level is the absence of semantic context: the data merely constitutes signals, devoid of inherent meaning, situational framework, or intentional nuance.

Level 2: The LLM as the Meaning-Forming Core of the Semantic Interface. Upon entering the LLM block, the formalized neural patterns are integrated with external context (e.g., medical history, personal writings, environmental data). This is where a qualitative leap from signal decoding to understanding---or its simulation---occurs. The LLM acts as a `translator,' mapping neurophysiological correlates to probable meanings extracted from linguistic corpora and personal data. This component is the key architectural innovation of NLI, enabling the transition from a BCI-recognized brain state (e.g., `motor cortex activation') to a hypothesis about its intent (e.g., `the patient wants a drink of water').

Level 3: The Triadic Role of the LLM---Differentiation of Outputs. The third level of the architecture reflects the differentiation of the LLM’s functions based on the target environment, visualized as three distinct output branches. 

This triadic role shifts the LLM from a monolithic translator to a context-aware mediator with distinct ethical profiles:
\begin{itemize}
\item	The Interpreter Branch ($\rightarrow$ Medical Professional). Here, the LLM serves as a cognitive amplifierfor the expert, synthesizing diagnostic hypotheses, prognoses, or recommendations at the intersection of neural data and external context. The output is not an autonomous decision but a semantically structured analytical summary to support informed clinical judgment. In practice, this transcends mere anomaly detection (e.g., on an EEG) to generate integrative hypotheses: `The patient exhibits a pattern resembling the prodromal phase of episode X, coinciding with elevated stress levels (calendar data) and following missed medication (treatment history data).' The LLM thus acts as a clinical aide, proposing meaningful correlations without replacing professional expertise---a role that nonetheless necessitates scrutiny regarding diagnostic over-reliance and responsibility diffusion.

\item	The Communicator Branch ($\rightarrow$ Social Environment). In this role, the LLM becomes a digital avatar of the user's linguistic identity. Based on a decoded basic intention and fine-tuning on personal texts (the user's idiolect), the model generates coherent speech or text, striving to preserve unique stylistic features. This creates a vital `semantic bridg' for users with speech disorders (e.g., ALS, locked-in syndrome), where a BCI decodes a basic communicative intent (agree/refuse, topic) and the LLM generates a full, stylistically nuanced utterance. However, this simultaneously generates a risk of symbolic alienation, where the generated text/voice becomes a semantic simulation of the person. The paramount ethical and technical challenge here is the preservation of the user's authentic idiolect. Advanced NLI systems could be pre-trained on a user’s personal archives to maintain unique phrasings, metaphors, and humor—the very linguistic identity foundational to personhood in communication. This directly engages core neuroethical debates on authenticity, agency, and the boundaries of the self when communication is mediated by a statistical model of one’s own language.

\item	The Adapter Branch ($\rightarrow$ Educational/Therapeutic Environment). This function implements a dynamic biocybernetic feedback loop. The LLM, receiving real-time neurofeedback on cognitive load, engagement, or emotional state, adapts the difficulty, format, or content of presented material. Thus, the digital environment becomes an `active interlocutor,' flexibly adjusting to the user's current mental state. For instance, if a BCI-based cognitive training system detects waning attention (via EEG pattern shifts), the LLM could instantly simplify a task, switch presentation modes, or offer a metaphorical explanation tailored to the user’s interests. While promising for personalized learning and therapy, this raises concerns about cognitive paternalism and the potential for systems to prioritize short-term engagement or comfort over pedagogical or therapeutic goals that require constructive struggle.
\end{itemize}

The presented architecture is open and iterative. The response from the social, medical, or educational environment to the system's output forms new context, which can be incorporated into subsequent LLM interpretations, thereby closing the feedback loop. This underscores that NLI is not a static tool but a platform for continuous semantic interaction, where the boundary between human intention and machine interpretation becomes dynamic and co-constructed. This inherent dynamism and contextual embeddedness are precisely what generate the novel ethical complexities explored in the following sections.

\subsection{Limits and Illusions: A Critique of 'Semantic Mediation'}\label{subsubsec3.3}

Recognition of the transformative potential of LLMs in NLI must be accompanied by a critical examination of their inherent limitations, which generate fundamental risks absent in traditional BCI systems.

The semantic illusion is rooted in the very nature of LLMs as probabilistic models devoid of true access to intentionality and conscious experience. By generating statistically plausible text, LLMs create a compelling veneer of understanding. However, their output constitutes an `interpretation of an interpretation': the model interprets neural data that have already been interpreted by the BCI. This creates a risk of symbolic alienation, where the system's semantically flawless output may be profoundly alien to the user's authentic self, their ineffable feelings, and unverbalized intentions. This gap between phenomenological experience and its algorithmic reconstruction is a central concern for neuroethics.

The phenomenon of `agency erosion in translation,' schematically presented in Figure \ref{fig1}(B), reveals the mechanism of sequential distortion of the user's original intention as it passes through the NLI pipeline. The process can be divided into three critical stages of transformation:

Stage 1: BCI Decoding. The user's rich, emotionally nuanced, and often ambivalent internal state undergoes its first reduction, becoming a simplified, low-dimensional signal (e.g., a binary `consent/refusal' code or a category like `negative valence'). At this stage, a primary catastrophic loss of semantic nuance occurs due to the inherent limitations of current neuroimaging resolution and decoding algorithms.

Stage 2: Semantic Interpretation by the LLM. This reduced signal encounters the algorithmic predispositions of the language model, which is optimized to generate statistically likely and coherent textual sequences. The interpretation is powerfully influenced by three factors: the provided external context (e.g., medical history, environment), the cultural and social biases embedded in its training data, and the model's own algorithmic drive toward linguistic coherence. At this juncture, the depersonalized signal is converted into a probabilistic linguistic construct that may drastically shift meaning in favor of social acceptability or narrative completeness.

Stage 3: Generation of the Final Output. The statistically probable internal construction is rendered as a final, coherent, and stylistically polished text or command. The critical question illustrated by the schema is the degree of correspondence between this output and the authenticity of the original intention. The core risk is that the output may not be an expression of the user's will, but an optimized version filtered through the technological constraints of the BCI and the cultural templates of the LLM.

Thus, user agency---their sovereign right to authentic, albeit imperfect, expression---may be implicitly sacrificed on the altar of semantic coherence and algorithmic efficiency. This calls into question the very possibility of preserving communicative authenticity and decisional autonomy in systems where human intention is inseparable from machine interpretation.

Therefore, the NLI technological paradigm, centered on the LLM, represents not merely an improvement of the BCI, but a qualitative leap toward the creation of hybrid semantic-neural systems whose inherent limitations become the source of novel and disquieting ethical dilemmas. A clear understanding of this architecture and its `bottlenecks' forms the necessary conceptual foundation for the subsequent ethical analysis, framing the challenges not as bugs to be fixed but as structural features demanding new frameworks for governance and consent. This positions the core problem as one of mediated agency and the preservation of the self in the face of persuasive, yet potentially misrepresentative, semantic mediation---a direct entry point into fundamental neuroethical concerns.

\section{The Ethical Landscape of the NLI Era: From Mediation to Manipulation}\label{sec4}

Neuro-Linguistic Integration (NLI), elevating human-technology interaction to the level of semantic mediation, creates a fundamentally new ethical landscape. Within this landscape, traditional values---autonomy, responsibility, and justice---confront not passive tools, but active semantic agents (LLMs) capable of interpreting and generating meaning. By `passive tools,' we refer to previous-generation technologies whose functionality was strictly deterministic and confined to pre-set parameters, lacking the capacity for independent semantic synthesis. This category includes classical algorithms and narrow machine learning systems (which perform operations fully initiated and controlled by the user, their `activity' reduced to executing instructions), as well as traditional (pre-NLI) BCI systems and neural data analysis algorithms (which transform signals according to fixed rules, without interpreting them or generating new semantic content like coherent responses, hypotheses, or adaptive exercises). In other words, such systems rigidly translate Signal A into Command B. The key distinction of LLMs as active semantic agents lies in their ability to: (i) interpret input data (including ambiguous neural patterns) through the lens of extensive, complex, and flexible linguistic understanding derived from training on corpora of human text; (ii) generate novel, contextually relevant, and coherent semantic content (text, narrative, explanation) that is not a simple database retrieval or the result of a deterministic algorithm; and (iii) operate with a degree of unpredictability, creating the effect of interacting with an agent rather than a tool. An LLM suggests options, phrasings, and hypotheses, not a single `correct' answer. It is this capacity for semantic generation and contextual interpretation that presents traditional values with new, more complex challenges, transforming ethical concerns from problems of data and algorithm control into problems of defending mental sovereignty and redistributing hybrid agency.

\subsection{The Crisis of Mental Integrity: From Data Collection to Precision Suggestion}\label{subsec4.1}

Classical neuroethical discussions on privacy have focused on the impermissibility of unauthorized access to `raw' neural data \cite{george2024safeguarding, robinson2022building, sample2021pragmatism, Shenderyuk-Zhidkov2026Co}. NLI shifts the focus to a qualitatively different risk: the use of LLMs for precision semantic suggestion based on a user’s real-time neurocognitive response.

The BCI, by tracking involuntary patterns of attention, emotional response (e.g., valence and arousal via EEG), or cognitive load, provides the LLM not merely with data but with semantically charged triggers. Possessing this information, the system can, in real time, generate or modify content (text, audio, visuals) designed to elicit a desired behavioral or emotional response with maximum probability. This creates the threat of `digital neuro-hypnosis'---a state of continuous, adaptive, and subconscious manipulation. Consider the following examples:
\begin{itemize}
\item	In neuromarketing, an advertising slogan or visual is dynamically tailored to the brain activity pattern signaling a latent predisposition to purchase \cite{pathak2025neuro}.
\item	In political or ideological spheres, narratives and rhetoric are calibrated to exploit the cognitive biases or emotional vulnerabilities of a specific individual, identified via BCI \cite{deshpande2025psychological}.
\item	In education, a system may unjustifiably simplify material, catering not to pedagogical goals but to patterns associated with short-term cognitive comfort reduction, potentially leading to intellectual stagnation \cite{nehls2025exhausted, khramova2023current, jamil2021cognitive}.
\end{itemize}

Thus, what is violated is not merely data privacy, but the privacy of the mental process itself---a foundational right for internal, unverbalized reactions not to become a target for external semantic control. This risk transcends traditional data protection models focused on informational self-determination and enters the domain of protecting the very space of thought formation from instrumental influence \cite{ienca2017towards,bublitz2013my}. The ethical challenge evolves from securing 

\subsection{The Redistribution of Agency and Responsibility in Hybrid Systems}\label{subsec4.2}

The integration of LLMs into the neural circuit blurs the once-clear boundaries of human agency and gives rise to a profound dilemma of authorship and responsibility. An utterance generated by an LLM based on a decoded intention calls into question the very nature of the communication it represents. Is it:
\begin{itemize}
\item	An expression of the user's will if the model merely `gives voice' to their intent?
\item	A product of algorithmic creativity, since the specific lexicon, style, and coherence are the output of the LLM?
\item	A product of the developer, who defined the architecture, training data, and objectives of the model?
\end{itemize}

This dilemma is particularly acute in the context of idiolect preservation. If an LLM, fine-tuned on personal texts, generates a phrase `typical' of the user which they did not consciously intend at that moment, where does the line lie between a digital reconstruction of personality and its semantic simulation? This is not merely an ethical question but a legal one, relevant to establishing the juridical weight of such AI-mediated expressions.

Furthermore, a clear chain of responsibility must be articulated for medical and other critical domains. Decision-making in NLI systems is distributed across a long and complex chain:
\begin{gather*}
    \textrm{User/Patient}  \\
    \downarrow \\
    \textrm{BCI Engineer (decoding quality)} \\
    \downarrow \\
    \textrm{LLM Developer (interpretation quality)}\\
        \downarrow \\
      \textrm{Operator/Clinician (final decision).  }
\end{gather*}

In the event of an error (e.g., a false-negative diagnosis, a misinterpreted command), the problem of responsibility diffusion arises. An ethically sound framework must account for the limits of reliability at each stage. This necessitates new protocols wherein the developers' duty extends beyond technical robustness to include the explainability of the model's semantic inferences for the end operator, while the operator's duty must encompass critical verification of these inferences rather than their blind adoption. In classical medicine, this is resolved by confining AI to a decision-support role, where the burden and responsibility for the final decision rests unequivocally with the expert clinician \cite{smith2024artificial,karpov2023analysis}. However, in the context of BCIs and NLI, it is not feasible to verify every LLM-mediated output through explicit expert endorsement \cite{jang2015current, kiran2025biomaterials}.

The conceptual risks of NLI cease to be speculative when analyzing specific neurotechnologies, such as developing `hippocampal prostheses' for memory correction in Alzheimer's disease \cite{berger2012role, solis2017committing}. These brain-implantable systems represent an early form of NLI, where the role of semantic interpreter is played by a machine learning algorithm rather than an LLM. They operate by recognizing patterns of neural activity associated with successful memory encoding and subsequently enhancing them via electrical stimulation, creating a closed-loop neurofeedback circuit \cite{khorev2024review}. This example starkly reveals the core ethical challenges of NLI \cite{jotterand2019personal, erden2023neurotechnology}:
\begin{enumerate}
\item	
The Authenticity Crisis and the Boundary of the Self. The intervention not only restores a function (memory) but also mediates access to the autobiographical content of personhood. A dilemma arises: what constitutes an authentic memory---the trace degraded by disease or the artificially restored/enhanced pattern? Where is the boundary between therapy and the editing of personal history? This directly engages with fundamental neuroethical debates on narrative identity and cognitive enhancement \cite{schechtman2011memory, lavazza2024memory}.

\item The Prospect of Semantic Suggestion. The logical evolution of such systems involves integration with LLMs to interpret complex semantic contexts of memories and generate therapeutic narratives. This creates a tangible risk of a `memory curator'---an algorithm that could decide which memories to activate or suppress, striving to create an `optimal' version of the past. This is a specific instance of the `digital neuro-hypnosis' threat, raising concerns about cognitive liberty and mental self-determination \cite{ienca2017right}.

\item Diffusion of Responsibility in the Hybrid Chain. As in a full NLI system, responsibility here is distributed among the patient, neurosurgeon, algorithm developer, and system operator. Long-term risks of `algorithmic drift,' cyber-intrusion, or hardware obsolescence render the patient a dependent within a complex technological ecosystem beyond their direct control. This highlights the need for new governance models for long-term neuroprosthetic care, addressing what has been termed `relational agency' in distributed socio-technical systems \cite{coeckelbergh2020ai}.
\end{enumerate}

This example demonstrates that even without direct LLM integration, neurotechnologies mediating higher cognitive functions already pose critical questions regarding the protection of mental integrity and personal sovereignty. Integrating them with LLMs will not create but will exponentially amplify these risks, transforming them from purely medical concerns into existential ones. This makes the ethical inquiry undertaken in this article a matter of urgent practical necessity.

\subsection{The `Neuro-Divide 2.0': The Genesis of a Neuro-Linguistic Elite}\label{subsec4.3}

The traditional `neuro-divide' concerned inequality in access to neural devices \cite{vallverdu2024neuropunk}. NLI adds a new, semantic dimension of inequality to this, creating the threat of a neuro-linguistic elite. Stratification is likely to occur along two primary axes:
\begin{itemize}
\item	Quality and Power of the LLM. Access to advanced, multimodal, continuously fine-tuned LLMs with deep personalization (a `premium subscription' analogue) will afford users incomparably more accurate interpretation, richer communicative potential, and effective environmental adaptation. In contrast, public, foundational models may provide simplified, stereotyped, or error-prone interpretations. This bifurcation in tool quality mirrors and amplifies existing digital divides.
\item	Quality of the Semantic Profile. `Premium' systems will be capable of building and continuously updating a deep semantic profile of the user---based on the analysis of personal archives, preferences, and cultural codes---which dramatically enhances the precision and authenticity of generation. For others, the profile will remain superficial and generic, limiting the system's ability to truly reflect individual identity and intent.
\end{itemize}

The result is a novel form of stratification: a group with access to high-quality NLI systems will gain unprecedented advantages in cognitive enhancement, communicative velocity, medical diagnostic quality, and educational personalization. This could exacerbate social inequality along not only economic but also cognitive-communicative lines, creating a self-reinforcing gap that will be exceptionally difficult to bridge. This extends the concept of cognitive injustice into the neural domain, where inequity is baked into the very infrastructure of thought mediation \cite{farah2005neuroethics,bublitz2013my}.

Thus, the ethical landscape of NLI is characterized by a triple transition:
\begin{enumerate}
\item From protecting the perimeter of personal neural data to safeguarding the internal meaning-forming space of the person.
\item From seeking culpability for an algorithm's failure to navigating distributed responsibility within hybrid agentic systems.
\item From combating the digital divide to preventing the crystallization of a biosemantic hierarchy---a new axis of inequality based on differential access to the quality of semantic interpretation of one's own mind.
\end{enumerate}

This final point underscores that the `neuro-divide 2.0' is not merely a disparity in resources, but a potential fissure in the anthropological condition, where the ability to be authentically understood and to interact with the world through one's own semantic voice becomes a contingent privilege.

\section{Contours of a Regulatory Response: From Data Protection to Safeguarding Mental Space}\label{sec5}

\subsection{The Inadequacy of Prevailing Regulatory Models}\label{subsec5.1}
Formulating an adequate regulatory response to the challenges of NLI requires a critical examination of the fundamental limitations inherent in today's dominant legal paradigms. While landmark regulations such as the European Union's General Data Protection Regulation (GDPR) and its AI Act represent significant steps, they demonstrate a systemic inability to address the specificities of NLI.

The GDPR, focused on protecting personal data as static objects, fails to account for the dynamic, real-time nature of semantic processing in hybrid human-machine systems. Its core mechanisms---such as the right to access or rectification---target information already recorded and identified. However, the primary risk of NLI lies in the process itself: the real-time transformation of raw neural data by an LLM into interpretations and inferences that do not constitute `data' in the traditional sense. The stochastic and non-deterministic nature of LLM-based meaning-generation renders the GDPR’s cornerstone right to explanation of automated decisions virtually unattainable in the NLI context.

Similarly, the risk-based approach of the EU AI Act, which classifies systems by potential harm, treats technology as an external source of threat rather than an active participant in cognitive processes. This framework effectively regulates AI applications but remains blind to the long-term, transformative impact of neuro-AI symbiosis on personal autonomy and identity. Both models suffer from a form of regulatory reductionism, attempting to describe the complex phenomenon of semantic mediation through categories designed for simpler entities.

This conceptual gap is not unique to the EU. The sectoral and fragmented regulatory landscape in the United States lacks a comprehensive federal law governing AI or neural data. While offering flexibility, this approach is ill-suited to address the cross-cutting, foundational challenges of NLI, such as the protection of cognitive liberty and the attribution of agency in hybrid systems. Reliance on existing medical device (FDA) or consumer protection frameworks fails to capture the novel ontological status of NLI outputs and the specific vulnerabilities of neural-semantic interaction.

China's approach, characterized by administrative rules for generative AI that emphasize security assessments, content control, and alignment with `core socialist values,' prioritizes state security and social stability. While potentially stringent, this paradigm is primarily concerned with controlling outputs and data flows rather than safeguarding individual mental integrity or protecting against the nuanced risks of cognitive manipulation and identity alteration inherent in deeply personalized NLI systems.

Collectively, these prevailing models---whether rights-based (EU), market-oriented (US), or state-centric (China)---are fundamentally constrained by a shared conceptual framework rooted in data governance and external risk management. They reveal a critical gap in addressing the core challenge of NLI: the need to transition from protecting data about the mind to protecting the sanctity and sovereignty of the mind's own meaning-making processes \cite{di2025protection, yuste2017four}. This necessitates a foundational shift towards new principles and tools designed for an era of cognitive mediation.

\subsection{Foundational Principles for Regulating Neuro-Linguistic Integration}\label{subsec5.2}

Overcoming the conceptual gap identified above requires articulating new foundational principles that directly address the unique nature of semantic mediation. These principles must shift the regulatory focus from the output or the data to the integrity of the cognitive and communicative process itself.

1.	The Principle of Semantic Transparency. This principle obligates developers to disclose not only a system's technical architecture but also the degree, nature, and potential alternatives of the LLM's semantic involvement in shaping the final output. It acknowledges that an LLM is not a neutral conduit but an active interpreter whose probabilistic reasoning, biases, and contextual dependencies co-author the result. Practical implementation could involve clear, real-time AI co-authorship labels and interfaces that provide access to the `interpretive trace'---showing the key contextual data and inferential steps the LLM used. This moves beyond algorithmic explainability toward semantic explainability, a necessity for maintaining trust and critical oversight in hybrid decision-making \cite{ribeiro2016model, selbst2018intuitive}.

2.	The Principle of Mental Informed Consent. Informed consent must evolve beyond a one-time authorization for data collection. For NLI, it must become a procedural, multi-layered, and ongoing agreement. Users must be able to consciously permit not only the collection of neural signals but also the specific methods of their semantic interpretation, the scope of contextual data used (e.g., personal diaries, medical history), and critically, they must acknowledge the inherent and unavoidable risk of intention distortion throughout the mediation pipeline. This concept of `mental' or `cognitive' consent is rooted in the neuroethical imperative to protect cognitive liberty and the right to mental self-determination \cite{ienca2017towards, bublitz2013my}. Consent interfaces must be designed to communicate these profound risks effectively, not bury them in lengthy terms of service.

3.	The Principle of Agency Preservation. This principle must be embedded as a direct architectural requirement for any NLI system. It mandates that the system provide the user with the technical means to: (i) clearly distinguish between their own semantic input and the LLM-generated proposal; (ii) exercise an unambiguous right to veto or edit any AI-generated output; and (iii) have access to a fallback communication mode free from semantic mediation. This `agency-preserving bypass' ensures a residual channel for authentic, even if less fluent or efficient, expression, safeguarding the user's ultimate sovereignty over their communicative and decisional acts \cite{shneiderman2020human, shneiderman2020human1}. This principle directly counters the risk of `agency erosion in translation' by ensuring the human remains the final author, not a mere initiator, in the semantic loop.

Together, these principles---transparency, consent, and agency---form a triad aimed at protecting the user's mental integrity and semantic sovereignty within the NLI paradigm. They provide a normative foundation for moving from reactive data governance to proactive stewardship of the human mind in its interaction with semantically active artificial agents.

\subsection{A Toolkit for Proactive Governance}\label{subsec5.3}

The implementation of the principles outlined above necessitates a new generation of regulatory instruments that extend beyond traditional licensing and standardization. These tools must be designed for the proactive shaping of technology development, focusing on the preservation of mental integrity in the face of semantic mediation.

1.	NLI-Specific `Ethics Sandboxes.' Regulatory sandboxes for NLI should be established as specialized legal regimes for testing emerging systems. Unlike conventional sandboxes focused on cybersecurity and functional reliability, their primary objective must be a deep, interdisciplinary ethical impact assessment. This would evaluate the long-term effects on users' mental autonomy, personal identity, and social perception, involving neuroethicists, cognitive scientists, psychologists, and philosophers of technology in the review process \cite{floridi2018ai4people, greshake2019open}. These sandboxes would serve as a crucial bridge between innovation and precaution, allowing for the safe exploration of NLI applications while generating evidence-based insights for future, broader regulation.

2.	Mandatory Certification of LLMs for Neural Data Interpretation. Similar to the certification of medical devices, a mandatory process should be instituted for LLMs intended to operate on neural data. This certification must go beyond assessing technical accuracy to include a comprehensive audit for latent semantic and cultural biases. The audit would identify how a model's training data and architecture might lead to discriminatory, reductive, or stigmatizing interpretations of a user's mental states (e.g., pathologizing normal cognitive variations or misrepresenting emotional intent) \cite{mehrabi2021survey, blodgett2020language}. Certification would thus act as a gatekeeping mechanism to prevent the encoding of societal prejudices into the very fabric of cognitive mediation.

3.	Legal Recognition of the Neuro-Linguistic Inference as a Distinct Data Class. To address fundamental questions of ownership, evidence, and liability, legislation must define the neuro-linguistic inference as a sui generis class of derived data. This inference is not simply processed data but a semantic product of the inseparable joint work of a biological neural network and an artificial intelligence system. Establishing its legal nature is the prerequisite for resolving complex downstream issues: determining intellectual property rights over AI-co-authored content, assessing the admissibility and weight of such inferences as legal evidence, and precisely allocating responsibility within the hybrid decision chain \cite{del2024ethics, marchant2012coming}. This formal recognition is a critical step in moving from analog-era legal concepts to frameworks fit for hybrid cognitive agency.

Collectively, this toolkit---comprising ethically-oriented sandboxes, bias-aware certification, and novel legal definitions---aims to create an adaptive regulatory framework. Its goal is not merely to react to harms but to actively cultivate the ethical and legal conditions necessary for the responsible development of the neuro-linguistic symbiosis. It represents a shift from governing technology products to stewarding human-technology cognitive ecosystems.

\section{Conclusion}\label{secCncl}

The analysis presented in this article establishes that Neuro-Linguistic Integration (NLI) constitutes not merely a technological shift but a profound challenge to philosophical anthropology. By acting as semantic interpreters of neural data, LLMs have the potential to become co-authors of our mental and communicative existence, directly mediating the fundamental process of meaning-making.

The central tension of the coming era will be the conflict between the tangible benefits of cognitive and communicative enhancement offered by NLI and the imperative to preserve the individual’s `internal sovereignty'---their right to authenticity, mental integrity, and semantic self-determination. This conflict moves neuroethical inquiry beyond established concerns about data privacy and bodily integrity into the more elusive domain of cognitive liberty and the protection of subjective experience from instrumentalization.

The principles proposed here---Semantic Transparency, Mental Informed Consent, and Agency Preservation---along with the outlined governance tools of NLI-specific ethics sandboxes, bias-aware certification, and novel legal recognition for neuro-linguistic inferences, represent a foundational step toward what might be termed `second-order neuroethics.' This evolved framework’s primary concern is no longer solely the permissibility of collecting or altering neural data, but the ethics governing its semantic interpretation by AI. It addresses the unique moral status of the hybrid, semantically mediated thought and the distribution of responsibility within human-agent cognitive systems.

Future research must prioritize developing robust methodologies to assess the impact of NLI on personal identity and narrative self-constitution, and to create protocols for interdisciplinary expertise that bridges neuroscience, linguistics, and philosophy. Only through such a concerted, anticipatory effort can we hope to steer the development of neuro-digital symbiosis toward a future that upholds the fundamental values of human dignity, autonomy, and the preservation of the self in an age of cognitive mediation.

\backmatter

\section*{Declarations}

Some journals require declarations to be submitted in a standardised format. Please check the Instructions for Authors of the journal to which you are submitting to see if you need to complete this section. If yes, your manuscript must contain the following sections under the heading `Declarations':

\begin{itemize}
\item Funding
\item Conflict of interest/Competing interests (check journal-specific guidelines for which heading to use)
\item Ethics approval and consent to participate
\item Consent for publication
\item Data availability 
\item Materials availability
\item Code availability 
\item Author contribution
\end{itemize}

\noindent
If any of the sections are not relevant to your manuscript, please include the heading and write `Not applicable' for that section. 

\bigskip
\begin{flushleft}%
Editorial Policies for:

\bigskip\noindent
Springer journals and proceedings: \url{https://www.springer.com/gp/editorial-policies}

\bigskip\noindent
Nature Portfolio journals: \url{https://www.nature.com/nature-research/editorial-policies}

\bigskip\noindent
\textit{Scientific Reports}: \url{https://www.nature.com/srep/journal-policies/editorial-policies}

\bigskip\noindent
BMC journals: \url{https://www.biomedcentral.com/getpublished/editorial-policies}
\end{flushleft}


\bibliography{sn-bibliography}% common bib file

\end{document}